\documentclass[a4paper,notitlepage]{article}

\usepackage[utf8]{inputenc}
\usepackage[T1]{fontenc}

\usepackage{amsxtra, amsfonts, amssymb, amstext}
\usepackage{amsthm}
\usepackage{longtable}
\usepackage{xspace}
\usepackage{lscape}
\usepackage[algo2e,ruled,vlined,linesnumbered]{algorithm2e}
\newcommand{\assign}{\leftarrow}
\usepackage{graphicx}

\usepackage{caption}
\usepackage{subcaption}

\usepackage{enumitem}
\setitemize{noitemsep,topsep=0pt,parsep=0pt,partopsep=0pt,leftmargin=*}

\usepackage[affil-it]{authblk}
\usepackage{hyperref}

\newtheorem{theorem}{Theorem}

\newcommand{\ignore}[1]{}

\allowdisplaybreaks[1]

\newcommand{\N}{\mathbb{N}}
\newcommand{\R}{\mathbb{R}}

\renewcommand{\epsilon}{\varepsilon}

\DeclareSymbolFont{bbold}{U}{bbold}{m}{n}
\DeclareSymbolFontAlphabet{\mathbbold}{bbold}

\DeclareMathOperator{\mut}{flip}

\DeclareMathOperator{\avg}{avg}

\DeclareMathOperator{\Bin}{Bin}

\newcommand{\onemax}{\textsc{OneMax}\xspace}

\newcommand{\OM}{\textsc{Om}\xspace}

\newcommand{\leadingones}{\textsc{LeadingOnes}\xspace}
\newcommand{\LO}{\textsc{Lo}\xspace}
\newcommand{\jump}{\textsc{Jump}\xspace}

\newcommand{\oea}{$(1 + 1)$~EA\xspace}

\newcommand{\oeares}{$(1 + 1)$~EA$_{>0}$\xspace}

\newcommand{\opl}{$(1+\lambda)$~EA\xspace}
\newcommand{\opladap}{$(1+\lambda)$~EA$_{r/2,2r}$\xspace}

\newcommand{\oplres}{$(1+\lambda)$~EA$_{>0}$\xspace}

\begin{document}
\title{Towards a Theory-Guided Benchmarking Suite for Discrete Black-Box Optimization Heuristics:\\ Profiling $(1+\lambda)$ EA Variants on OneMax and LeadingOnes} 

\author[1]{Carola Doerr}
\affil[1]{Sorbonne Universit\'e, CNRS, LIP6, Paris, France}

\author[2]{Furong Ye}

\author[2]{Sander van Rijn}

\author[2]{Hao Wang}

\author[2]{Thomas B\"ack}
\affil[2]{LIACS, Leiden University, Leiden, The Netherlands}

\date{July 17, 2018}


\maketitle
\begin{abstract}
Theoretical and empirical research on evolutionary computation methods complement each other by providing two fundamentally different approaches towards a better understanding of black-box optimization heuristics. In discrete optimization, both streams developed rather independently of each other, but we observe today an increasing interest in reconciling these two sub-branches. In continuous optimization, the COCO (COmparing Continuous Optimisers) benchmarking suite has established itself as an important platform that theoreticians and practitioners use to exchange research ideas and questions. No widely accepted equivalent exists in the research domain of discrete black-box optimization.

Marking an important step towards filling this gap, we adjust the COCO software to pseudo-Boolean optimization problems, and obtain from this a benchmarking environment that allows a fine-grained empirical analysis of discrete black-box heuristics. In this documentation we demonstrate how this test bed can be used to profile the performance of evolutionary algorithms. More concretely, we study the optimization behavior of several $(1+\lambda)$~EA variants on the two benchmark problems OneMax and LeadingOnes. This comparison motivates a refined analysis for the optimization time of the $(1+\lambda)$~EA on LeadingOnes. 
\end{abstract}





\section{Introduction}\label{sec:intro}

After a long phase of around 20 years in which mathematical and empirical research on evolutionary algorithms (EAs) seemed to have advanced more or less independently of each other, we observe today a steadily growing interest to reconcile these two different methodologies. This interest is spurred by the awareness that certain advantages of one approach cannot be compensated for by the other: While theoretical work has the advantage of universal bounds that are proven with mathematical rigor, empirical research contributes concrete numbers (as opposed to the predominant asymptotic bounds derived by theoretical approaches) and can provide results for complex problems or algorithms that do not admit a stringent mathematical analysis. This way, both research methods offer a different view on the working principles of evolutionary computation (EC) methods. Bringing these two streams closer together has the potential to create a much more complete picture than what any of the two approaches can deliver individually.  

In continuous optimization, a strong promoter for joint projects and discussion is the \href{http://coco.gforge.inria.fr/}{COCO (COmparing Continuous Optimisers) platform}, which offers a well-designed and widely accepted benchmarking environment for derivative-free continuous black-box optimizers. The affiliated BBOB workshops have established themselves as a prime platform for theoreticians and practitioners to exchange novel ideas and research questions. For discrete black-box optimization, unfortunately, no initiative of equivalent reputation exists. 

With this work we wish to make an important step towards establishing a well-designed benchmarking environment for pseudo-Boolean black-box optimization. To this end, we build on the freely available COCO software~\cite{COCO} (cf.~\cite{hansen2016cocoplat} for a documentation of this framework) and adapt it to discrete benchmark problems. We summarize in this work how this framework can be used to generate novel and---as we shall see---sometimes quite surprising findings that, in turn, inspire new research directions. 

To demonstrate the functioning of the benchmarking platform, we investigate in this work the performance of different \opl variants. These variants differ in their parameter settings, i.e., in the offspring population size $\lambda$ and the mutation rates $p$ used by the variation operator. Apart from \opl variants with static parameter settings, we also investigate the performance profiles of four different algorithms with self-adjusting parameter choices. 

To provide a comparison that is meaningful for theoreticians and practitioners alike, we respond to a recent critique of the way EAs are typically evaluated in the theory of EC literature~\cite{CarvalhoD17} and regard in our comparison \opl variants that do not create offspring that are identical to their parents. For our static and noiseless benchmark problems, such offspring would not provide any new information to the algorithm, and since a plus selection scheme is in place, they do also not bring any other benefit to the optimization process. In our \oplres variants, we therefore ensure that every offspring differs from the parent in at least one bit. 

\subsection{The Role of Benchmarking in EC}
Benchmarking aims at supporting practitioners in choosing appropriate algorithms for a given optimization problem through a systematic empirical comparison of different algorithmic techniques. Typical research questions that can be addressed by a benchmarking approach concern 
\begin{itemize}
	\item the \emph{scalability} of performance in terms of problem dimension or decision variables,  
	\item the optimization behavior across different \emph{benchmark functions}, 
	\item the dependence of performance on selected \emph{problem and instance features}, and 
	\item the \emph{sensitivity} of a heuristic with respect to small changes of its parameters.
\end{itemize}

For theoreticians, benchmarking can be an essential tool for the enhancement of mathematically-derived ideas into techniques that can be beneficial for a broad range of problems. It can also be used to identify reasonable parameter settings for which performance bounds can be proven. Benchmarking is also an important tool to communicate mathematical research results, as it can be more appealing to look at performance plots than to compare mathematical formulae. Finally, benchmarking can also serve as a source for interesting research questions. We present such an example in Section~\ref{sec:LOnewbound}, where we prove refined runtime bounds for the \opl on the \leadingones problem. This result is clearly motivated by the observation that the $(1+1)$~EA$_{>0}$ and the $(1+50)$~EA$_{>0}$ show a very similar performance in our empirical comparisons.

\subsection{Performance Criteria}
Most theoretical works use \emph{total expected optimization time}, i.e., the average number of function evaluations needed to identify an optimal solution, as the only performance criterion. This measure is very coarse, as it reduces the whole optimization process to a single number. It has therefore been suggested in~\cite{JansenZ14} to complement optimization time with a more fine-grained performance measure. Jansen and Zarges~\cite{JansenZ14} suggested to adapt a \emph{fixed-budget perspective,} and to measure the expected quality of a solution that can be obtained within a given computational budget. A complementary performance measure, which is more commonly used for algorithm comparison, are \emph{fixed-target results,} which measure the expected number of function evaluations needed to identify solutions of a certain target quality. This perspective extends optimization time to arbitrary target values. Our benchmarking tool, just like COCO, provides both these performance measures.

\subsection{Comment on the Benchmark Problems}
At the current state, our benchmarking environment is set up for the comparison of black-box optimizers on a few selected benchmark problems, such as linear functions, functions with a plateau (so-called jump-functions), or the non-separable \leadingones functions. This selection is clearly biased towards functions that are analyzable by mathematical means. We will extend this selection significantly in the near future. 

We have chosen to present in this work a comparison of performance profiles for the two functions \onemax and \leadingones. For these two problems a number of theoretical results exist, and one may have the feeling that these problems are quite well understood already. Our empirical comparison shows nevertheless some surprising effects, which have inspired us to prove a more precise runtime bound for the performance of the \opl on \leadingones. The empirical performance profiles also raise a number of other interesting questions that may be useful to address by a mathematical approach. 


We recall that theoreticians regard simple benchmark functions like \onemax and \leadingones in the hope that, among other reasons,
\begin{itemize}
	\item they give insights into how the studied algorithms perform on the easier parts of a difficult optimization problem,
	\item in order to understand some basic working principles of the algorithms, which can then be used for the analysis of more complex problems, more complex algorithms, and for the development of new algorithmic ideas, 
	\item the theoretical investigations (which even for seemingly simple algorithms and problems can be surprisingly complex) triggers the development of new analytical tools that can be used for more complex tasks, and
	\item very precise mathematical statements can be obtained, which allow to determine, for example, how the chosen parameter values influence the performance of an algorithm, or how these parameters can be controlled in an optimal way.
\end{itemize}

We furthermore note that even for \onemax and \leadingones, despite being studied since the very early days of theory of evolutionary computation, several unsolved problems exist, many of which are of seemingly simple nature such as the optimal dynamic mutation rate of the \oea for \onemax. We are confident that our benchmarking environment will be a key enabler to strengthen existing results, to guide researchers towards interesting research questions, and, ultimately, towards high-performing optimization techniques. 

\section{Summary of Algorithms}
\label{sec:algos}

Two fundamental building blocks of evolutionary algorithms are global variation operators and populations. \emph{Global variation operators} are sampling strategies that are characterized by the property that every possible solution candidate has a positive probability of being sampled within a short time window, regardless of the current state of the algorithm. Standard bit mutation is an example of a global mutation operator. From a given input string $x\in\{0,1\}^n$, standard bit mutation creates an offspring by flipping each bit in $x$ with some positive probability $0<p<1$, with independent decisions for each bit. For any $x,y$ the probability to sample $y$ from $x$ is thus $p^{H(x,y)}(1-p)^{n-H(x,y)}$, where $H(x,y)$ is the number of bits in which $x$ and $y$ differ (\emph{Hamming distance}). This probability is positive even for search points that are very far apart. The motivation to use global sampling strategies is to overcome local optima by eventually performing a sufficiently large jump. 

Storing information about the optimization process, maintaining a diverse set of reasonably good solutions, and gathering a more complete picture about the structure of the problem at hand are among the most important reasons to employ \emph{population-based EAs.} The first two objectives are served by the \emph{parent population}, which is the subset of previously evaluated search points that are kept in the memory of the algorithm. The parent population is updated after each generation. New solution candidates are sampled from it through the use of variation operators. These points form the \emph{offspring population} of the generation. Non-trivial offspring population sizes address the desire to gather more information about the fitness landscape before making any decision about which of the points from the parent and offspring population to keep in the memory for the next iteration. 

It is very well understood that both the size of the parent population as well as the size of the offspring population can have a significant impact on the performance. Finding suitable parameter values for these two quantities remains to be a challenging problem in practical applications of EAs. From an analytical point of view, populations increase the complexity of the optimization process considerably, as they introduce a lot of dependencies that need to be taken care of in the mathematical analysis. It is therefore not surprising that only few theoretical works on population-based EAs exist, cf.~\cite{LehreO17tut} and references mentioned therein. Most existing theoretical works regard algorithms with non-trivial \emph{offspring population} sizes, while the impact of the \emph{parent population} size has received much less interest. 

Since one of the goals of this work is to showcase how benchmarking can serve as a platform for theoreticians and practitioners to discuss new research directions, we present below empirical and theoretical results for the \opl, the arguably simplest EA that combines a global sampling technique with a non-trivial offspring population size. The \opl and its various variants analyzed in Sections~\ref{sec:OM} and~\ref{sec:LO} are formally defined below. Existing theoretical results for the selected benchmark problems will be presented in the respective sections. 

\textbf{Notation.} Throughout this work, the problem dimension is denoted by $n$. As common in evolutionary computation, we assume that the algorithms know the dimension of the problem that they are optimizing. We further assume that we aim to \emph{maximize} the objective function. For every positive integer $r$, we abbreviate by $[r]$ the set $\{1,2,\ldots,r\}$ and we set $[0..r]:=\{0\} \cup [r]$. By $\ln$ we denote the natural logarithm to base $e:=\exp(1)$.

\subsection{Motivation for the Modifications}\label{sec:motimodi}
As noted in~\cite{CarvalhoD17} there exists an important discrepancy between the algorithms classically regarded in the theory of evolutionary computation literature and their common implementations in practice. For mutation-based algorithms like $(\mu+\lambda)$ and $(\mu,\lambda)$~EAs, this discrepancy concerns the way new solution candidates are sampled from previously evaluated ones, and how the function evaluations are counted. Both algorithms use the above-described standard bit mutation as only variation operator. An often recommended value for the mutation rate $p$ is $1/n$, which corresponds to flipping exactly one bit on average, an often desirable behavior when the search converges. 

When implementing standard bit mutation, it would be rather inefficient to decide for each $i \in [n]$ whether or not the $i$-th bit of $x$ should be flipped. Luckily, this is not needed, as we can simply observe that standard bit mutation can be equally expressed as drawing a random number $\ell$ from the binomial distribution $\Bin(n,p)$ with $n$ trials and success probability $p$ and then flipping $\ell$ bits that are sampled from $[n]$ uniformly at random (u.a.r.) and without replacement. This latter operation is formalized by the $\mut_{\ell}$ operator in Algorithm~\ref{alg:mut}. We refer to $\ell$ as the \emph{mutation strength} or the \emph{step size}, while we call $p$ the \emph{mutation rate}.  

\begin{algorithm2e}[t]%
	\textbf{Input:} $x=(x_1 \ldots x_n) \in \{0,1\}^n$, $\ell \in \N$\;
		  $y \assign x$\;
		\label{line:elloea}Select $\ell$ pairwise different positions $i_1,\ldots,i_{\ell} \in [n]$ u.a.r.\;
		\lFor{$j=1,...,\ell$}{$y_{i_j}\assign 1-x_{i_j}$}
\caption{$\mut_{\ell}$ chooses $\ell$ different positions and flips the entries in these positions.}
\label{alg:mut}
\end{algorithm2e}

Analyzing standard bit mutation, we easily observe that the probability to not flip any bit at all equals $(1-p)^n$, which for $p=1/n$ converges to $1/\exp(1)$. That is, in about $36.8\%$ of calls to this operator, a copy of the input is returned. For the \opl there is no benefit of evaluating such a copy (unless facing a dynamic or noisy optimization setting), since it applies \emph{plus selection}, where both the parent as well as the offspring can be selected to ``survive'' for the next generation. It is therefore advisable to change the probability distribution from which the mutation strength $\ell$ is sampled. A straightforward (and commonly used) idea is to simply re-sample $\ell$ from $\Bin(n,p)$ until a non-zero value is returned. This approach corresponds to distributing the probability mass $(1-p)^n$ of sampling a zero proportionally to all step sizes $\ell>0$. This gives the conditional binomial distribution $\Bin_{>0}(n,p)$, which assigns to each $\ell \in \N$ a probability of $\binom{n}{\ell} p^{\ell} (1-p)^{n-\ell}/(1-(1-p)^n)$. All our empirical results use this conditional sampling strategy. The results can therefore differ significantly from figures previously published in the theory of EA literature.\footnote{We note that the suggestion to adapt the performance evaluation has been made several times in the literature, e.g., ~\cite{HemertB02,JansenZ11,CarvalhoD17}, but has not been picked up systematically.} 
%

\subsection{The Basic \texorpdfstring{$(1+\lambda)$ EA$_{>0}$}{(1+l) EA}}
The \opl samples $\lambda$ offspring in every iteration, from which only the best one survives (ties broken uniformly at random). Each offspring is created by standard bit mutation. Following our discussion above, we make use of the re-sampling strategy described above, and obtain the \oplres, which we summarize in Algorithm~\ref{alg:oplres}. 

 \begin{algorithm2e}%
	\textbf{Initialization:} 
	Sample $x \in \{0,1\}^{n}$ u.a.r.\;
  \textbf{Optimization:}
	\For{$t=1,2,3,\ldots$}{
		\For{$i=1,\ldots,\lambda$}{
			Sample $\ell^{(i)}$ from $\Bin_{>0}(n,p)$\;
			$y^{(i)} \assign \mut_{\ell^{(i)}}(x)$\;
		}
		\label{line:selectionopl}
		Sample $x$ from $\arg\max\{f(x),f(y^{(1)}), \ldots, f(y^{(\lambda)})\}$ u.a.r.\;	
	}
\caption{The \oplres with mutation rate $p \in (0,1)$ for the maximization of $f:\{0,1\}^n \rightarrow \R$}
\label{alg:oplres}
\end{algorithm2e}

\subsection{Adaptive \texorpdfstring{$(1+\lambda)$ EA$_{>0}$}{(1+l) EA} Variants}
The \oplres has two parameters: the offspring population size $\lambda$ and the mutation rate $p$. Common implementations of the \oplres use the same population size $\lambda$ and the same mutation rate $p$ throughout the whole optimization process (\emph{static parameter choice}), while the use of \emph{dynamic parameter values} is much less established. A few works exist, nevertheless, that propose to \emph{control} the parameters of the \opl online~\cite{KarafotiasHE15}. We focus in our empirical comparison on algorithms that have a mathematical support. These are summarized in the following two subsections. 

\subsubsection{Adaptive Mutation Rates}\label{sec:adaptivemut}

One of the few works that experiments with a \emph{non-static mutation rate} for the \opl was presented at GECCO~2017~\cite{DoerrGWY17}. The there-suggested algorithm stores a parameter $r$ that is adjusted online. In each iteration, the \opladap creates $\lambda/2$ offspring by standard bit mutation with mutation rate $r/(2n)$, and it creates $\lambda/2$ offspring with mutation rate $2r/n$. The value of $r$ is updated after each iteration. With probability $1/2$ it is set to the value that the best offspring individual of the last iteration has been created with (ties broken at random), and it is replaced by either $r/2$ or $2r$ otherwise (unbiased random decision). Finally, the value $r$ is capped at $2$ if smaller, and at $n/4$, if it exceeds this value. In our experiments, we use $r=2$ as initial value.

In~\cite{DoerrGWY17} it is shown analytically that the \opladap yields an asymptotically optimal runtime on \onemax. This performance is strictly better than what any static mutation rate can achieve, cf. Section~\ref{sec:OMtheo}. How well the adaptive scheme works for other benchmark problems is left as an open question in~\cite{DoerrGWY17}. 


\subsubsection{Adaptive Population Sizes}\label{sec:adaptivepop}
Apart from the mutation rate, one can also consider to \emph{adjust the offspring population size $\lambda$.} This is a much more prominent problem, because $\lambda$ is an explicit parameter, while the mutation rate is often not specified (and thus by default assumed to be $1/n$). 

In the theory of EC literature, the following three success-based update rules have been studied. In~\cite{JansenJW05} the offspring population size $\lambda$ is initialized as one. After each iteration, we count the number $s$ of offspring that are at least as good as the parent. When $s=0$, we double the population size, and we replace it by $\lfloor \lambda/s \rfloor$ otherwise. For brevity, we call this algorithm the $(1+\{2\lambda,\lambda/s\})$~EA, and its resampling variant the $(1+\{2\lambda,\lambda/s\})$~EA$_{>0}$. 

Two similar schemes were studied in~\cite{LassigS11} where $\lambda$ is doubled if no strictly better search point has been identified and either set to one or to $\max\{1,\lfloor \lambda/2 \rfloor\}$ otherwise. We regard here the resampling variants of these algorithms, which we call the $(1+\{2\lambda,1\})$~EA$_{>0}$ and the $(1+\{2\lambda,\lambda/2\})$~EA$_{>0}$, respectively.


\section{OneMax}\label{sec:OM}

We start our empirical investigation with the class of \onemax functions, the generalization of the function $\OM$ that assigns to each bit string $x$ the number $|\{ i\in [n] \mid x_i=1\}|$ of ones in it. For this generalization, $\OM$ is composed with all possible XOR operations on the hypercube. More precisely, for any bit string $z \in \{0,1\}^n$ we define the function $\OM_z:\{0,1\}^n \to [0..n], x \mapsto |\{ i\in [n] \mid x_i=z_i\}|$, the number of bits in which $x$ and $z$ agree. The \onemax problem is the collection of all functions $\OM_z$, $z \in \{0,1\}^n$.

To identify interesting parameter ranges for the offspring population size $\lambda$ and the mutation rate $p$, we first summarize known theoretical results in Section~\ref{sec:OMtheo}. Results of the performance profiling will be discussed in Section~\ref{sec:OMempi}. 

\subsection{Theoretical Bounds}\label{sec:OMtheo}

\onemax is often referred to as the \emph{drosophila of EC.} It is therefore not surprising that among all benchmark functions, \onemax is the problem for which most runtime results are available. Due to the space limit, we can only summarize a few selected results. 

Concerning the \opl, the first question that one might ask is whether or not it can be beneficial to generate more than one offspring per iteration. When using the number of function evaluations (and not the number of generations) as performance measure, intuitively, it should always be better to create the offspring sequentially, to profit from intermediate fitness gains. This intuition has been formally proven in~\cite{JansenJW05}, where it is shown that for all $\lambda,k \in \N$ the expected optimization time (i.e., the number of function evaluations until an optimal solution is queried for the first time) of the $(1+k\lambda)$~EA cannot be better than that of the \opl. This result implies that $\lambda=1$ is an optimal choice. 
 Note, however, that the number of \emph{generations} needed to find an optimal solution can significantly decrease with increasing $\lambda$, so that it can be beneficial even for \onemax to run the \opl with $\lambda>1$ when a parallel function evaluations are possible. 

For $\lambda=1$ the runtime of the \oea with \textbf{static mutation rate} $p>0$ is quite well understood, cf.~\cite{Witt13j} for a detailed discussion. For general $\lambda$ and static mutation rate $p=c/n$ (where here and henceforth $c>0$ is assumed to be constant), the expected optimization time is $(1\pm o(1)) \big( \frac{n \lambda \ln \ln \lambda}{2 \ln \lambda} + \frac{e^c}{c} n \ln n\big)$~\cite{GiessenW17Algorithmica}. An interesting observation made in~\cite{DoerrGWY17} reveals that the parametrization $p=c/n$ is suboptimal: with $p=\ln(\lambda)/(2n)$ the \opl needs only an expected number of $O(n\lambda / \log(\lambda) + \sqrt{\lambda} n \log n)$ function evaluations to optimize \onemax~\cite{DoerrGWY17} (the proof requires $\lambda\ge 45$ and $\lambda=n^{O(1)}$). We call this algorithm the $(1+\lambda)$~EA$_{p=\ln(\lambda)/(2n)}$. 

When using \textbf{non-static mutation rates}, the best expected optimization time that a \opl can achieve on \onemax is bounded from below by $\Omega(n\lambda / \log(\lambda) + n \log n)$~\cite{BadkobehLS14}. This bound is attained by a \opl variant with fitness-dependent mutation rates~\cite{BadkobehLS14}. Interestingly, it is also achieved by the self-adjusting \opladap described in Section~\ref{sec:adaptivemut} (the proof requires again $\lambda\ge 45$ and $\lambda=n^{O(1)}$).

Concerning the algorithms using \textbf{non-static offspring population sizes} (cf. Section~\ref{sec:adaptivepop}), we do not have an explicit theoretical analysis for the $(1+\{2\lambda,\lambda/s\})$~EA, but it is known that the expected optimization time of both the $(1+\{2\lambda,1\})$~EA and the $(1+\{2\lambda,\lambda/2\})$~EA is $O(n \log n)$~\cite{LassigS11}.

\textbf{Disclaimer.} 
It is important to note that all the bounds reported above (and those mentioned in Section~\ref{sec:LOtheo}) hold, a priori, only for the classical \opl variants, not the resampling versions regarded here in this work. For most bounds, and in particular the ones with static parameter choices, it is, however, not difficult to prove that the modifications do not change the asymptotic order of the expected optimization times. What does change, however, is the leading constant. As a rule of thumb, runtime bounds for the \opl decrease by a multiplicative factor of about $1-(1-p)^n$ when the mutation strengths are sampled from the conditional binomial distribution $\Bin_{>0}(n,p)$. 

Note also that in our summary we collect only statements about the total expected \emph{optimization time}. Fixed-target results for non-optimal target values are not yet very common in theoretical works on EAs, and neither are fixed-budget results. Here again the \oea and the \opl with static parameters form an exception as for these two algorithms a few theoretical fixed-budget results exist, cf.~\cite{LenglerS15,DoerrDY16,CorusHJOSZ17,NallaperumaNS17} and references therein.

\subsection{Empirical Evaluation}\label{sec:OMempi}

We now come to the results of our empirical investigation. All figures presented in this work are averages over 100 independent runs. This might look like a small number, but we recall that we track the whole optimization process, to present the fixed-target results below. As a rule of thumb, storing the data for the 100 runs of one algorithm on one dimension requires around 5-10 MB of storage. Despite the seemingly small number of repetitions, the numbers and results presented below are quite consistent. We also recall that in the COCO framework, an often recommended choice is to have around 15 independent runs, cf.~\cite[Section~3.1]{HansenABTT16}. 

We also note that for this work we have not yet adjusted the post-processing part of the COCO environment, so that the plots in this submission have been created by Microsoft Excel. We are working on making an automated post-processing available as soon as possible.  

\begin{figure}
\centering
\includegraphics[width=\linewidth]{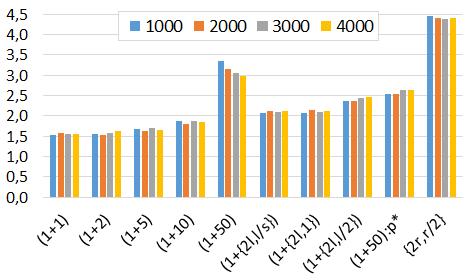}
\caption{Average optimization times for $100$ independent runs of the \oplres variants, normalized by $n\ln n$. (Initial) population size for the adaptive variants is $50$. $p^*=\ln(\lambda)/(2n)$}
\label{fig:OM-per-dim}
\end{figure}

We have seen above that, for reasonable parameter settings, the average optimization times of the \opl variants are all of order $n \log n$. We therefore normalize the empirical averages in Figure~\ref{fig:OM-per-dim} by this factor. 

We observe that the normalized averages are quite stable across the dimensions, with an exception of the $(1+50)$~EA, whose relative performance improves with increasing problem dimension. We also see that the $(1+50)$~EA$_{>0}$ variant with $p^*=\ln(\lambda)/(2n)$ achieves a better optimization time than the $(1+50)$~EA$_{>0}$ with $p=1/n$. The variants with adaptive offspring population size perform significantly worse than the \oeares, which is not surprising given the performance hierarchy of the $(1+\lambda)$~EAs mentioned in the beginning of Section~\ref{sec:OMtheo}.  

For the tested problem dimensions, the worst-performing algorithm in our comparison is the \opladap. This may come as a surprise since this algorithm is the one with the best theoretical support. The advantage of this algorithm seems to require much larger problem dimensions, different values of $\lambda$, and/or different settings of the hyper-parameters that determined its update mechanism. 

A general question raised by the data in Figure~\ref{fig:OM-per-dim} concerns the sensitivity of the adaptive \opl variants with respect to the initialization of their parameters and with respect to their hyper-parameters, which are the update strengths, but also the initial parameter values of $\lambda$ and $r$, respectively. 
%

\begin{figure*}
        \centering
        \begin{subfigure}{.5\textwidth}
            \centering
                \includegraphics[width=.9\textwidth]{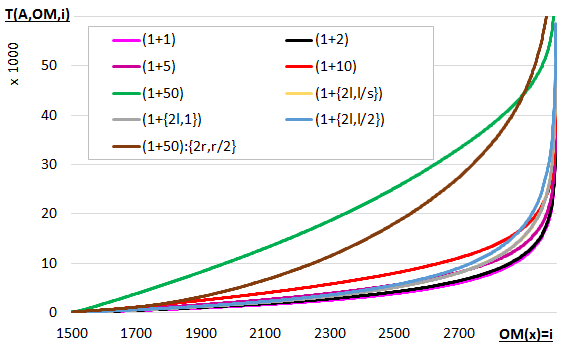}
                \caption{Average fixed-target runtimes $T(A,\OM,i)$. The difference of the $(1+\{2\lambda,\lambda/s\})$~EA$_{>0}$ to the $(1+\{2\lambda,1\})$~EA$_{>0}$ is less than $2\%$ for $\OM(x)$-values $\ge 1,800$, and the curves are therefore indistinguishable in this range.}
								\label{fig:OM-time-per-fx}
        \end{subfigure}
        ~
        \begin{subfigure}{.5\textwidth}
            \vspace{-4em}
            \centering
                \includegraphics[width=.9\textwidth]{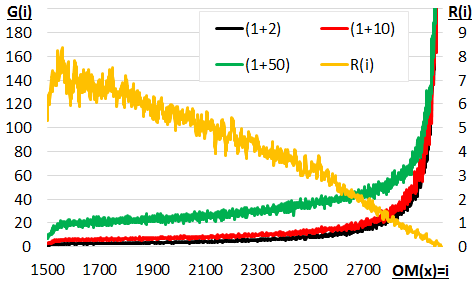}
                \caption{Gradients $G(A,\OM,i)$ and relative disadvantage $R(i)$ of the (1+50) EA$_{>0}$ over the (1+2) EA$_{>0}$}
								\label{fig:OM-time-per-progress}
        \end{subfigure}
\caption{Fixed target data for the $3,000$-dimensional \onemax problem (averages over 100 runs)}
\label{fig:gradientOM}
\end{figure*}

It has been discussed in Section~\ref{sec:OMtheo} that for \onemax the performance of the \opl can only be worse than that of the \oea. In fact, the data in Figure~\ref{fig:OM-per-dim} demonstrates a quite significant discrepancy between the performance of the \oeares and the \oplres variants with $\lambda \ge 10$. The expected optimization time of the $(1+50)$~EA$_{>0}$, for example, is about twice as large as that of the \oeares. Intuitively, this can be explained as follows. At the beginning of the \onemax optimization process the probability that a random offspring created by the \opl improves upon its parent is constant. In this phase the \opl variants with small $\lambda$ have an advantage as they can (almost) instantly make use of this progress, while the $(1+\lambda)$~EAs with large $\lambda$ first need to wait for all $\lambda$ offspring to be evaluated. Since the expected fitness gain of the best of these $\lambda$ offspring is not much larger than that of a random individual, large offspring population sizes are detrimental in this first phase of the optimization process. We note, however, that the relative disadvantage of large $\lambda$ is much smaller towards the end of the optimization process. When, say, the parent individual $x$ satisfies $\OM(x)=n-\Theta(1)$, the probability that a random offspring has a better function value is of order $\Theta(1/n)$ only. We therefore have to create $\Theta(n)$ offspring, in expectation, before we see any progress. The relative disadvantage of creating several offspring in one generation is therefore almost negligible in the later parts of the optimization process (provided that $\lambda=O(n)$). This informal explanation is confirmed by the plots in Figure~\ref{fig:gradientOM}, which display for $n=3,000$ 
\begin{enumerate}
	\item in Figure~\ref{fig:OM-time-per-fx}: the empirical average fixed-target times $T(A,\OM,i)$; i.e., the average number of function evaluations that the \oplres variant $A$ needs to identify a solution of $\OM$-value at least $i$. We cap this plot at $60,000$ evaluations.
	\item in Figure~\ref{fig:OM-time-per-progress}: the gradients $G(A,i):=T(A,\OM,i)-T(A,\OM,i-1)$ (three lowermost curves), and the relative difference $R_i:=(\avg_{j=i,i+1,...,i+5}(G((1+50) \text{EA}_{>0},j) - \avg_{j=i,i+1,...,i+5} G((1+2) \text{EA}_{>0},j)/G((1+2) \text{EA}_{>0},j))$ of the rolling average of the gradients of the $(1+50)$ and the $(1+2)$~EA$_{>0}$.
\end{enumerate}
Note that the gradient $G(A,i)$ measures the average time needed by algorithm $A$ to make a progress of one when starting in a point $x$ of \onemax-value $\OM(x)=i$. Small gradients are therefore desirable. We see that, for example, the $(1+50)$~EA$_{>0}$ (green curve) needs, on average, about 20 fitness evaluations to generate a strictly better search point when starting in a solution of $\OM$-value around 1,700. The $(1+2)$~EA$_{>0}$, in contrast, needs only about $2.6$ function evaluations, on average. The relative disadvantage of the $(1+50)$~EA$_{>0}$ over the $(1+2)$~EA$_{>0}$ decreases with increasing function values from around 7 to zero, cf. the uppermost (yellow) curve in Figure~\ref{fig:OM-time-per-progress}. We use the rolling average of 5 consecutive values here to obtain a smoother curve for $R(i)$.

Another important insight from Figure~\ref{fig:OM-time-per-fx} is that the dominance of the $(1+1)$~EA$_{>0}$ over all $(1+\lambda)$~EA$_{>0}$ variants does not only apply to the total expected optimization time, but also for to all intermediate target values. This can be shown with mathematical rigor by adjusting the proofs in~\cite[Section~3]{JansenJW05} to suboptimal target values.
\section{LeadingOnes}\label{sec:LO}

In this section we consider the performance of the \opl on \leadingones, which will shed a different light on the usefulness of large population sizes. Known theoretical bounds are summarized in Section~\ref{sec:LOtheo}. In Section~\ref{sec:LOempi} we discuss selected findings of our empirical investigation. These results inspire us to present a refined mathematical analysis for the expected optimization time of the \opl on \leadingones in Section~\ref{sec:LOnewbound}.  

Before presenting our results, we recall that \leadingones is the generalization of the function $\LO$, which counts the number of initial ones in the string, i.e., $\LO(x)=\max \{ i \in [0..n] \mid \forall j \leq i: x_j=1\}$. The generalization is by composing with an XOR-shift and a permutation of the positions. This way, we obtain for every $z \in \{0,1\}^n$ and for every permutation (one-to-one map) $\sigma$ of the set $[n]$ the function $\LO_{z,\sigma}:\{0,1\}^n \rightarrow \N$, which assigns to each $x$ the function value 
$
\max \{ i \in [0..n] \mid \forall j \in [i]:  x_{\sigma(j)} = z_{\sigma(j)} \}.
$ 
The \leadingones problem is the collection of all these functions. 

\subsection{Theoretical Bounds}\label{sec:LOtheo}
Also for \leadingones it has been proven that the optimal value of the offspring population size $\lambda$ in the \opl is one when using function evaluations and not generations as performance indicator~\cite{JansenJW05}. In contrast to \onemax, however, we will observe, by empirical and mathematical means, that the disadvantage of non-trivial population sizes is much less pronounced for this problem. 

The \oea with fixed mutation probability $p$ has an expected optimization time of $\frac{1}{2p^2}((1-p)^{-n+1}-(1-p))+1$ on \leadingones~\cite{BottcherDN10}, which is minimized for $p\approx 1.59/n$. This choice gives an expected runtime of about $0.77n^2$. A fitness-dependent mutation rate can decrease this runtime further to around $0.68 n^2$~\cite{BottcherDN10}. For the \oeares, it has been observed in~\cite{JansenZ11} that its expected optimization time decreases with decreasing $p$. More precisely, it equals $\frac{1-(1-p)^n}{2p^2}((1-p)^{-n+1}-(1-p))+1$, which converges to $n^2/2 + 1$ for $p \to 0$.

For $\lambda=n^{O(1)}$, the expected optimization time of the \opl with mutation rate $p=1/n$ is $O(n^2 + n\lambda)$~\cite{JansenJW05}. With this mutation rate, the adaptive $(1+\{2\lambda,\lfloor \lambda/2 \rfloor\})$~EA and the $(1+\{2\lambda,1\})$~EA achieve an expected optimization time of $O(n^2)$~\cite{LassigS11}. Any \opl variant with fixed offspring population size $\lambda$ but possibly adaptive mutation rate $p$ needs at least $\Omega(\frac{\lambda n}{\ln(\lambda/n)}+n^2)$ function evaluations, on average, to optimize \leadingones~\cite{BadkobehLS14}. 

As mentioned, we will revisit and refine the bound for the \opl in Section~\ref{sec:LOnewbound} but we first present the empirical results that have motivated this analysis. 

\subsection{Empirical Results}\label{sec:LOempi}

\begin{figure}
\centering
\includegraphics[width=\linewidth]{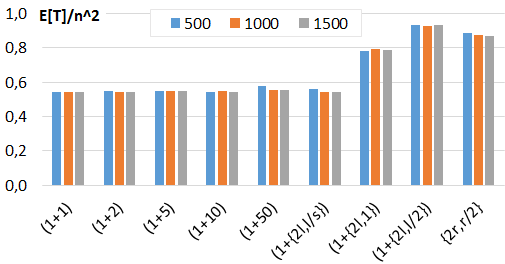}
\caption{Average optimization times for $100$ independent runs, normalized by $n^2$. (Initial) population size for the adaptive variants is $50$.}
\label{fig:LO-per-dim}
\end{figure}

Similarly to the data plotted in Figure~\ref{fig:OM-per-dim}, we show in Figure~\ref{fig:LO-per-dim} the normalized average optimization times of the different algorithms; the normalization factor is $n^2$. 

For all \opl variants, a fairly stable performance across the three tested dimensions $n=500$, $n=1,000$, and $n=1,500$ can be observed. 
We also see that the $(1+\{2\lambda,\lambda/s\})$~EA$_{>0}$ performs very well; it is just slightly worse than the (1+2) EA$_{>0}$. The $(1+\{2\lambda,\lambda/2\})$~EA$_{>0}$, the $(1+\{2\lambda,1\})$~EA$_{>0}$, and the \opladap, in contrast, perform significantly worse than any of the tested \oplres. For the former two algorithms, preliminary test show that the disadvantage can be decreased by adjusting the update rule to the one used by the $(1+\{2\lambda,\lambda/s\})$~EA$_{>0}$, i.e., to base the update decision upon the presence of offspring that are at least as good as the parent. With this update rule the performances become very similar to that of the $(1+\{2\lambda,\lambda/s\})$~EA$_{>0}$. The \opladap will be discussed in more detail below. 

Another interesting observation is that the value of $\lambda$ does not seem to have a significant impact on the expected performance. This is in sharp contrast to the situation for \onemax, cf. our discussion in the second half of Section~\ref{sec:OMempi}. Building on our discussion there, we can explain this phenomenon as follows. Unlike for \onemax, the situation for \leadingones is that the expected fitness gain of a random offspring created by a \opl variant with static mutation rate $p=c/n$ is very small throughout the whole optimization process. More precisely, it decreases only mildly from around $2c/n$ when $\LO(x)=0$ to $2c(1-c/n)^{n-1}/n \approx 2c/(e^{c}n)$ for $\LO(x)=n-1$. Thus, intuitively, the whole optimization process of \leadingones is very similar to the last steps of the \onemax optimization. 

\begin{figure}[t]
\centering
\includegraphics[width=0.9\linewidth]{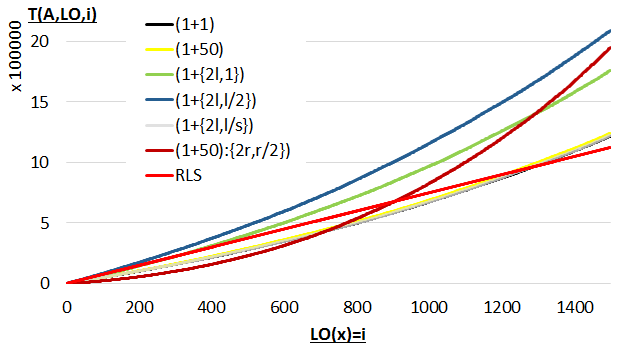}
\caption{Fixed-target runtimes of the \oplres variants on the $1,500$-dimensional \leadingones problem}
\label{fig:LO-time-per-fx}
\end{figure}

Figure~\ref{fig:LO-time-per-fx} presents the average fixed-target runtimes for selected \oplres variants on the $1,500$-dimensional \leadingones problem. We add to this figure the fixed target runtime of Randomized Local Search (RLS), the greedy (1+1)-type hill climber that always flips one random bit per iteration. RLS has a constant expected fitness gain of $2/n$ on \leadingones and thus a total expected optimization time of $n^2/2+1$.

We observe that the performance of the \oeares, the $(1+50)$~EA$_{>0}$, and the $(1+\{2\lambda,\lambda/s\})$~EA$_{>0}$ are indeed very similar throughout the optimization process, the curves are almost indistinguishable. The same applies to the \oplres with $\lambda=2,5,10$, these algorithms are therefore not shown in Figure~\ref{fig:LO-time-per-fx}. We also see that the fixed-target performance of these algorithms is better than that of RLS for all \LO-values up to around $1,250 \approx 0.42 n$ (exact empirical values are $1,227$ for the $(1+50)$~EA$_{>0}$ and $1,283$ for the \oeares, but we recall that such numbers should be taken with care as they represent an average of 100 runs only. It should not be very difficult to compute the cutting point precisely, by mathematical means). 

Since the only difference between the \oeares and RLS is the distribution from which new offspring are sampled, we see that the \opl variants profit from iterations in which more than one bit are flipped in the beginning of the optimization process, while they suffer from this same effect in the later parts. This situation is even more pronounced in the \opladap, which outperforms all other tested algorithms for target values up to $709$. Its performance then suffers from creating at least half of its offspring with a too large mutation rate. Recall that even if the algorithm has correctly identified the optimal mutation rate $p(\LO(x))$, it still creates half of its offspring with mutation rate $2p(\LO(x))$. 
This results in a mediocre overall performance. This observation certainly raises the question of how to adjust the structure of the \opladap to benefit from its good initial performance. From the viewpoint of hyper-heuristics, or algorithm selection, an adaptive selection between the \opladap and the \oplres would be desirable. 

If we had looked only at the total optimization times, we would have classified the \opladap as being inefficient. The fixed-target results, however, nicely demonstrate that despite the poor overall performance, there is something to be learned from this algorithm. This emphasizes the need for a fine-grained benchmarking environment.

\subsection{Precise Bounds for LeadingOnes}\label{sec:LOnewbound}

We have observed that, for \leadingones, the runtimes of the \oplres variants with static parameter choices are very close to that of the \oeares. In Theorem~\ref{thm:LOtheo} we show that for every constant $\lambda$, the expected optimization time of the \oplres converges from above against that of the \oeares(and the same holds for the \opl and \oea, respectively). Theorem~\ref{thm:LOtheo} can be proven by adjusting the proofs in~\cite{BottcherDN10} to the \opl. An important ingredient in this analysis is the observation that the probability of making progress in one \emph{generation} with parent individual $x$ equals $1-(1-p(1-p)^{\LO(x)})^{\lambda}$, i.e., 1 minus the probability that none of the $\lambda$ offspring is better. Recall that in order to create a better offspring, none of the first $\LO(x)$ bits should flip, while the $(\LO(x)+1)$-st bit \emph{does} need to be flipped. The results for the \oplres can be obtained from that for the \opl by taking into account that the re-sampling strategy increases the expected fitness gain by a multiplicative factor of $1/(1-(1-p)^n)$. 

\begin{theorem}\label{thm:LOtheo}
For all $n,\lambda \in \N$ the expected optimization time of the \opl with static mutation rate $0<p<1$ on the $n$-dimensional \leadingones function is at most 
\begin{align}\label{eq:oplLO}
1+\frac{\lambda}{2} \sum_{j=0}^{n-1}{ \frac{1}{1-(1-p(1-p)^j)^{\lambda}} }
\end{align}
and the expected optimization time of the \oplres is at most 
\begin{align}\label{eq:oplres}
1+\frac{(1-(1-p)^n)\lambda}{2} \sum_{j=0}^{n-1}{ \frac{1}{1-(1-p(1-p)^j)^{\lambda}} }.
\end{align}
\end{theorem}

To judge the precision of the bound stated in Theorem~\ref{thm:LOtheo}, we first note that for $\lambda=1$ expression~\eqref{eq:oplLO} is tight by the result presented in~\cite{BottcherDN10} (note though that the additive $+1$ term is suppressed there as they regard the number of iterations, not function evaluations). Similarly, for the \oeares expression~\eqref{eq:oplres} is tight by the bound proven in~\cite{JansenZ11}. 

Apart from this case with trivial offspring population size $\lambda=1$, it might be tedious to compute the expected optimization time of the \opl on \leadingones exactly, since in our proof for Theorem~\ref{thm:LOtheo} we would have to take into account that in one generation more than one search point that improves upon the current best search point can be generated. Since the \opl chooses the best one of these, the distribution of this offspring would have to be computed. Note, however, that this effect can only have a very mild impact on the bounds stated above, as is occurs relatively rarely and does, in general, not result in a much larger fitness gain. Put differently, the bounds in Theorem~\ref{thm:LOtheo} are close to tight for reasonable (i.e., not too large) values of $\lambda$.

As the expressions in Theorem~\ref{thm:LOtheo} are not easy to interpret, we provide in 
Table~\ref{tab:LOtheo} a numerical evaluation of the upper bound~\eqref{eq:oplres} for different values of $\lambda$ and $n$. We add to this table (second row) the empirically observed averages, which show a good match to the theoretical bound.

{\footnotesize{
	\begin{table}[h!]
\centering
\begin{tabular}{c|c|c|c|c|c|c}
\hline
	 &	 500   	&	 1,000   	&	 1,500     	&	 10,000   	&	 100,000   	&	 500,000   	\\
	\hline
(1+1) &	54.317\%	&	54.313\%	&	54.311\%		&	54.309\%	&	54.308\%	&	54.308\%	\\
emp.	&	54.0\%	&	54.1\%	&	54.2\%	    	&	 -     	&	 -     	&	 -     	\\
\hline															
(1+2) 	&	54.349\%	&	54.328\%	&	54.322\%		&	54.310\%	&	54.308\%	&	54.308\%	\\
 emp.	&	54.8\%	&	54.4\%	&	54.2\%	&	 -     	    	&	 -     	&	 -     	\\
\hline															
(1+5) 	&	54.444\%	&	54.376\%	&	54.353\%		&	54.315\%	&	54.309\%	&	54.308\%	\\
 emp.	&	54.5\%	&	54.8\%	&	54.6\%	&	 -          	&	 -     	&	 -     	\\
\hline															
(1+50) 	&	55.883\%	&	55.091\%	&	54.829\%	&	54.386\%	&	54.316\%	&	54.310\%	\\
 emp.	&	57.6\%	&	55.3\%	&	55.2\%	&	 -     	    	&	 -     	&	 -     	\\
\hline
\end{tabular}
\caption{Theoretical upper bounds from Theorem~\ref{thm:LOtheo}}
\label{tab:LOtheo}
\end{table}
}}

\vspace{-5ex}
\section{Conclusions}\label{sec:conclusions}

Building on the COCO software we have developed a benchmark environment that can be used to create fine-grained performance profiles for pseudo-Boolean benchmark problems. Results for \onemax and \leadingones have been presented in this documentation. These results inspired a refined analysis of the expected optimization time of the \opl on \leadingones. 

Our mid-term objective is to develop the benchmarking environment into a platform that practitioners and theoreticians use to exchange ideas and research questions on pseudo-Boolean black-box optimization. A key design question will be the selection of problems that shall be included in the benchmark suite. Our initial focus has been on functions for which some theoretical understanding of typical optimization processes is available, such as linear functions, \leadingones, \jump, etc. Going forward, we intend to include optimization problems that shed light on how the performance is influenced by certain problem features, such as the modality, the separability, the degree of constraints, etc.

\subsubsection*{Acknowledgments}
This work was supported by a public grant as part of the Investissement d'avenir project, reference ANR-11-LABX-0056-LMH, LabEx LMH, in a joint call with Gaspard Monge Program for optimization, operations research and their interactions with data sciences. Parts of our work have been inspired by discussions of 
COST Action CA15140 Improving Applicability of Nature-Inspired Optimisation by Joining Theory and Practice.

\bibliography{1Lambda-EA-Profiling-GECCO-18}

\begin{thebibliography}{10}

\bibitem{DoerrDY16}
{Benjamin Doerr}, {Carola Doerr}, and {Jing Yang}.
\newblock Optimal parameter choices via precise black-box analysis.
\newblock In {\em {GECCO'16}}, pages 1123--1130. {ACM}, 2016.

\bibitem{DoerrGWY17}
{Benjamin Doerr}, {Christian Gie{\ss}en}, {Carsten Witt}, and {Jing Yang}.
\newblock The $(1+\lambda)$~evolutionary algorithm with self-adjusting mutation
  rate.
\newblock In {\em {GECCO'17}}, pages 1351--1358. {ACM}, 2017.

\bibitem{Witt13j}
{Carsten Witt}.
\newblock Tight bounds on the optimization time of a randomized search
  heuristic on linear functions.
\newblock {\em {Combinatorics, Probability {\&} Computing}}, 22:294--318, 2013.

\bibitem{GiessenW17Algorithmica}
{Christian Gie{\ss}en} and {Carsten Witt}.
\newblock The interplay of population size and mutation probability in the
  $(1+\lambda)$ {EA} on {{O}ne{M}ax}.
\newblock {\em Algorithmica}, 78(2):587--609, 2017.

\bibitem{CorusHJOSZ17}
{Dogan Corus}, {Jun He}, {Thomas Jansen}, {Pietro~S. Oliveto}, {Dirk Sudholt},
  and {Christine Zarges}.
\newblock On easiest functions for mutation operators in bio-inspired
  optimisation.
\newblock {\em Algorithmica}, 78:714--740, 2017.

\bibitem{CarvalhoD17}
{Eduardo~Carvalho Pinto} and {Carola Doerr}.
\newblock Discussion of a more practice-aware runtime analysis for evolutionary
  algorithms.
\newblock In {\em {EA'17}}, pages 298--305, 2017.

\bibitem{KarafotiasHE15}
{G. Karafotias}, {M. Hoogendoorn}, and {A.E. Eiben}.
\newblock Parameter control in evolutionary algorithms: Trends and challenges.
\newblock {\em {IEEE Transactions on Evolutionary Computation}}, 19:167--187,
  2015.

\bibitem{BadkobehLS14}
{Golnaz Badkobeh}, {Per~Kristian Lehre}, and {Dirk Sudholt}.
\newblock Unbiased black-box complexity of parallel search.
\newblock In {\em {PPSN'14}}, volume 8672 of {\em {LNCS}}, pages 892--901.
  Springer, 2014.

\bibitem{HemertB02}
{Jano~I. van Hemert} and {Thomas B{\"{a}}ck}.
\newblock Measuring the searched space to guide efficiency: The principle and
  evidence on constraint satisfaction.
\newblock In {\em {PPSN'02}}, volume 2439 of {\em {LNCS}}, pages 23--32.
  Springer, 2002.

\bibitem{LenglerS15}
{Johannes Lengler} and {Nicholas Spooner}.
\newblock Fixed budget performance of the {(1+1)} {EA} on linear functions.
\newblock In {\em {FOGA'15}}, pages 52--61. {ACM}, 2015.

\bibitem{LassigS11}
{J{\"o}rg L{\"a}ssig} and {Dirk Sudholt}.
\newblock Adaptive population models for offspring populations and parallel
  evolutionary algorithms.
\newblock In {\em {FOGA'11}}, pages 181--192. {ACM}, 2011.

\bibitem{hansen2016cocoplat}
{N. Hansen}, {A. Auger}, {O. Mersmann}, {T. Tu{\v s}ar}, and {D. Brockhoff}.
\newblock {COCO}: A platform for comparing continuous optimizers in a black-box
  setting.
\newblock {\em {ArXiv e-prints}}, arXiv:1603.08785, 2016.

\bibitem{COCO}
{N. Hansen}, {A. Auger}, {O. Mersmann}, {T. Tu{\v s}ar}, and {D. Brockhoff}.
\newblock {COCO} github page.
\newblock \url{https://github.com/numbbo/coco}, [n. d.].

\bibitem{HansenABTT16}
{Nikolaus Hansen}, {Anne Auger}, {Dimo Brockhoff}, {Dejan Tusar}, and {Tea
  Tusar}.
\newblock {COCO:} performance assessment.
\newblock {\em {CoRR}}, abs/1605.03560, 2016.

\bibitem{LehreO17tut}
{Per~Kristian Lehre} and {Pietro~Simone Oliveto}.
\newblock Runtime analysis of population-based evolutionary algorithms:
  introductory tutorial at {GECCO} 2017.
\newblock In {\em {GECCO'17 Companion Material}}, pages 414--434. {ACM}, 2017.

\bibitem{NallaperumaNS17}
{Samadhi Nallaperuma}, {Frank Neumann}, and {Dirk Sudholt}.
\newblock Expected fitness gains of randomized search heuristics for the
  traveling salesperson problem.
\newblock {\em {Evolutionary Computation}}, 25, 2017.

\bibitem{BottcherDN10}
{S\"untje B{\"o}ttcher}, {Benjamin Doerr}, and {Frank Neumann}.
\newblock Optimal fixed and adaptive mutation rates for the {L}eading{O}nes
  problem.
\newblock In {\em {PPSN'10}}, volume 6238 of {\em {LNCS}}, pages 1--10.
  Springer, 2010.

\bibitem{JansenZ11}
{Thomas Jansen} and {Christine Zarges}.
\newblock Analysis of evolutionary algorithms: from computational complexity
  analysis to algorithm engineering.
\newblock In {\em {FOGA'11}}, pages 1--14. {ACM}, 2011.

\bibitem{JansenZ14}
{Thomas Jansen} and {Christine Zarges}.
\newblock Performance analysis of randomised search heuristics operating with a
  fixed budget.
\newblock {\em {TCS}}, 545:39--58, 2014.

\bibitem{JansenJW05}
{Thomas Jansen}, {Kenneth~A. {De Jong}}, and {Ingo Wegener}.
\newblock On the choice of the offspring population size in evolutionary
  algorithms.
\newblock {\em {Evolutionary Computation}}, 13:413--440, 2005.

\end{thebibliography}
\bibliographystyle{plain} 

\end{document}